\documentclass{article}

\PassOptionsToPackage{numbers, compress}{natbib}

\usepackage[preprint]{neurips_2022}

\usepackage[utf8]{inputenc} 
\usepackage[T1]{fontenc}    
\usepackage{hyperref}       
\usepackage{url}            
\usepackage{booktabs}       
\usepackage{amsfonts}       
\usepackage{nicefrac}       
\usepackage{microtype}      
\usepackage{xcolor}         
\usepackage{multirow, adjustbox}  
\usepackage{subcaption}     

\usepackage{geometry}
\usepackage{algorithm}
\usepackage{algpseudocode}

\algnewcommand{\IIf}[1]{\State\algorithmicif\ #1\ \algorithmicthen}
\algnewcommand{\EndIIf}{\unskip\ \algorithmicend\ \algorithmicif}
\newcommand{\SComment}[1]{\textcolor{blue}{\hfill/* \textit{#1} */}}

\usepackage{graphicx}

\usepackage{wrapfig}

\title{Active-Learning-as-a-Service: An Automatic and Efficient MLOps System for Data-Centric AI}

\author{%
      Yizheng Huang $^\ast$ \\
      Institute for Infocomm Research \\ A*STAR \\
      \texttt{huangyz0918@ieee.org} \\
    \And
    Huaizheng Zhang  \thanks{Equal contribution.}\\
    Nanyang Technological University \\
    \texttt{huaizhen001@e.ntu.edu.sg} \\
    \And
    Yuanming Li \\
    Institute of High Performance Computing \\ A*STAR \\
    \texttt{yuanmingleee@gmail.com} \\
    \And
    Chiew Tong Lau \\
    Nanyang Technological University \\
    \texttt{asctlau@ntu.edu.sg} \\
    \And
    Yang You \\
    National University of Singapore \\
    \texttt{youy@comp.nus.edu.sg}
}

\begin{document}

\maketitle

\begin{abstract}

The success of today's AI applications requires not only model training (Model-centric) but also data engineering (Data-centric). In data-centric AI, active learning (AL) plays a vital role, but current AL tools 1) require users to manually select AL strategies, and 2) can not perform AL tasks efficiently. To this end, this paper presents an automatic and efficient MLOps system for AL, named ALaaS (Active-Learning-as-a-Service). Specifically, 1) ALaaS implements an AL agent, including a performance predictor and a workflow controller, to decide the most suitable AL strategies given users' datasets and budgets. We call this a predictive-based successive halving early-stop (PSHEA) procedure. 2) ALaaS adopts a server-client architecture to support an AL pipeline and implements stage-level parallelism for high efficiency. Meanwhile, caching and batching techniques are employed to further accelerate the AL process. In addition to efficiency, ALaaS ensures accessibility with the help of the design philosophy of configuration-as-a-service. Extensive experiments show that ALaaS outperforms all other baselines in terms of latency and throughput. Also, guided by the AL agent, ALaaS can automatically select and run AL strategies for non-expert users under different datasets and budgets. Our code is available at \url{https://github.com/MLSysOps/Active-Learning-as-a-Service}.

\end{abstract}

\section{Introduction}
\label{Introduction}



Data-centric AI is an emerging topic that focuses on engineering data to develop AI applications with off-the-shelf machine learning (ML) models \cite{landingai}. Previous efforts are mainly model-centric AI that assumes a static environment. In this environment, 1) the data collection and engineering are done, 2) and continuously developing ML models to achieve high performance on test sets is the main target \cite{eyuboglu2022dcbench}. However, real-world AI applications are facing more complicated scenarios, which can not be adequately addressed by model-centric AI. For instance, researchers and practitioners have to spend a lot of time on data preparation, including data labeling \cite{chew2019smart}, and error detection \cite{krishnan2017boostclean}. Meanwhile, they also need to monitor the data, so that they can update models in time when a distribution drift is detected \cite{huang2021modelci}. Treating these issues only from a model-centric view will lead to a sub-optimal solution. Therefore, to further improve and democratize AI applications, a lot of efforts are now turning to data-centric or combining model-centric and data-centric \cite{landingai}.


Though the concept of data-centric AI has been proposed very recently, many pioneering studies whose core contributions lie in data engineering have already been proposed \cite{sener2018active, xu2021dataclue}. Among them, one vital direction is active learning (AL) \cite{ren2020survey}. The motivation of AL is to reduce manual labeling efforts while maintaining and even improving ML models' performance \cite{wang2014new, sener2018active, gal2017deep, ducoffe2018adversarial, caramalau2021sequential, ash2020deep, agarwal2020contextual, vzliobaite2013active, loy2012stream}. Specifically, it is well-known that ML models are very data-hungry. Therefore, to reach a high performance (\textit{e.g.}, accuracy) that meets application requirements, people always need to label a large amount of data. This process is extremely time-consuming and labor-intensive and thus often becomes the bottleneck of ML application development. To cope with the issue, AL selects the most representative yet diverse training samples from a large training data pool by utilizing AL strategies. Then, it only sends the selected samples to an oracle (\textit{e.g.}, human annotators) to label. Next, ML models will only be trained on these labeled sub-datasets. By doing so, we can still obtain an ML model with competitive performance but save labeling and training costs a lot.


However, utilizing AL is a non-trivial task. First, selecting a suitable AL strategy for a specific scenario given the budget and target accuracy is hard for both experts and non-experts. As a result, users have to run AL in a trial-and-error manner, resulting in huge time and monetary waste. Second, applying AL to AI application development is not simply searching for, selecting, or implementing AL strategies. Instead, users have to build a backend to run the AL pipeline, tailored for their applications in their environment (\textit{e.g.}, a private cluster and AWS). In other words, they need to undertake much repetitive engineering work with boilerplate code. Third, users have to consider the efficiency and cost issues, as AL often runs on a vast dataset, and some AL strategies (\textit{e.g.}, committee-based \cite{dagan1995committee, melville2004diverse}) require running more than one ML model for data selection. Under-consideration will result in a long process time and additional cost. Though several open-source AL tools \cite{modal, deepal, libact, alipy} lower the barrier of applying AL, and they can meet neither automation nor efficiency requirements.

\begin{figure*}[t]
  \centering
  \includegraphics[width=1.0\textwidth]{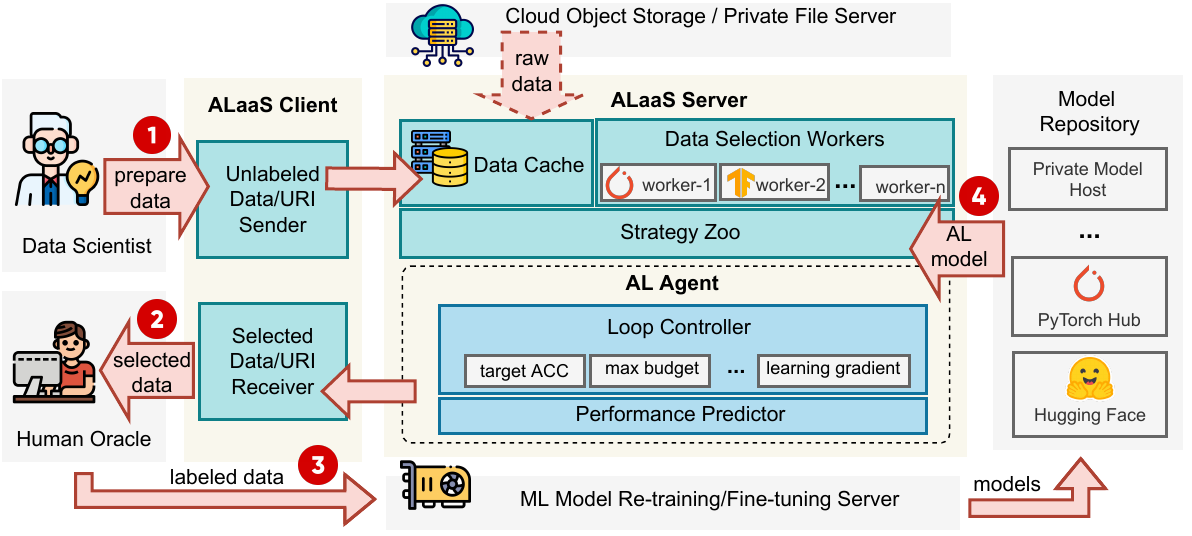}
  \caption{An overview of our ALaaS with human-in-the-loop. Our system adopts a server-client architecture. The data scientist first collects the unlabeled data and sends them to the AL server using an AL client, where an AL agent, a data cache and multiple workers will select data samples collaboratively and efficiently. The selected samples will be sent to a human oracle for labeling and then for updating models.}
\label{fig:system-arch}
\end{figure*}

To address these issues, we propose an automatic and efficient backend for AL. Our AL system, named Active-Learning-as-a-Service (ALaaS) (Figure \ref{fig:system-arch}). It can select AL strategies automatically given the budget and target accuracy without specifying an AL strategy. Also, it can run AL strategies on large datasets efficiently by utilizing single or distributed multiple devices. Specifically, ALaaS adopts the server-client architecture to perform AL tasks and can be deployed easily on both laptops and public clouds as a service. To achieve high efficiency, it implements a stage-level parallelism method to run AL tasks by fully utilizing hardware resources and reducing waiting time. Meanwhile, more acceleration techniques such as data cache and batching \cite{crankshaw2017clipper, zhang2020mlmodelci, zhang2020hysia} are utilized to further speed up the AL process. Besides, for users who have difficulty selecting a suitable AL strategy, ALaaS provides a predictive-based successive halving early-stop (PSHEA) procedure to automatically select and run AL with an only budget and target accuracy inputs. In addition to that, our system also considers accessibility and modularity, so that users can use many AL strategies in our AL zoo with ease, and experts can propose more advanced AL strategies for new scenarios. Experiments show that 1) our ALaaS outperforms all other baselines in terms of latency and throughput, and 2) the PSHEA procedure can predict future accuracy and select suitable AL strategies under different settings. Further ablation studies show the effectiveness of our design and reveal more insightful conclusions.





\section{Related Work}
This section presents the related work, including three categories: Active Learning (AL) algorithms and tools, Data-centric AI, and MLOps.

\subsection{AL Algorithms and Tools}

We categorize AL strategies into three classes, namely, diversity-based, uncertainty-based, and hybrid sampling. Diversity-based methods \cite{yang2015multi, sener2018active} are designed to select the most diverse samples from the whole dataset to represent it. Uncertainty-based methods \cite{wang2014new, roth2006margin, gal2017deep} aim to select the samples that can not be identified confidently by current ML models and then use these samples to further improve ML models. Hybrid methods \cite{huang2010active, beluch2018power} combine both the above-mentioned methods. Our system supports all of these methods and runs them more efficiently.

Many open-source AL tools have been developed to benefit both academia and industry, including ModAL\cite{modal}, DeepAL \cite{deepal}, Libact \cite{libact}, and ALiPy \cite{alipy}. Our ALaaS is inspired by these tools and further improves AL efficiency and accessibility by adopting the MLOps concept. The detailed comparison is summarized in Table\ref{tab:open-source-tool-compare}.

\subsection{Data-centric AI}

Data-centric AI is proposed to improve AI application performance by engineering datasets rather than only focusing on models. Recent Data-centric AI competitions and workshops \cite{landingai} demonstrate many exciting studies from academia and industry. Inspired by the pioneering work, many data-centric methods have been proposed for different areas, including Natural Language Processing (NLP) \cite{xu2021dataclue, seo2021automatic}, Computer Vision (CV) \cite{huang2021ymir, chakrabortyfirst}, Robotics \cite{lin2022roboflow}, etc. Also, a new benchmark \cite{eyuboglu2022dcbench} has been built for pushing forward data-centric AI research. To the best of our knowledge, ALaaS is the first MLOps system for efficient AL from the data-centric view.

\subsection{MLOps}

MLOps (Machine Learning Operation) aims to streamline the ML model development and reduce the AI application maintenance cost. Many MLOps systems have been proposed for both data-centric AI and model-centric AI \cite{https://doi.org/10.48550/arxiv.2102.07750}. From a data-centric view, labeling tools (\textit{e.g.}, labelme \cite{russell2008labelme}), data cleaning tools (\textit{e.g.}, ActiveClean \cite{krishnan2016activeclean}), data drift monitors and many others, can all be regarded as MLOps systems. From a model-centric view, there are model store systems \cite{vartak2016modeldb}, model continuous integration \cite{zhang2020mlmodelci, renggli2019continuous} tools, training platforms \cite{jiang2020unified}, and deployment platforms \cite{chen2018tvm}, etc. Different from these systems, ALaaS is designed specifically for executing human-involved AL tasks more efficiently.

In addition, tech giants starts to build end-to-end cloud platforms for MLOps (\textit{e.g.}, TFX \cite{baylor2017tfx}, SageMaker \cite{das2020amazon}, Ludwig \cite{molino2019ludwig}). Our ALaaS can be a good plugin complementary to these systems.

\begin{table}[t]
\centering
\caption{Comparison of AL open-source tools. Our ALaaS offers an automated Machine-Learning-as-a-Service experience and largely improves AL efficiency.}
\label{tab:open-source-tool-compare}
\vspace{7pt}
\centering
    \adjustbox{max width=\textwidth}{
    \begin{tabular}{lcccccc}
        \toprule
        \begin{tabular}[c]{@{}c@{}}AL \\Open-source Tool\end{tabular} & \begin{tabular}[c]{@{}c@{}}Pipelined \\Data Processing\end{tabular} & \begin{tabular}[c]{@{}c@{}}Automatically \\Strategy Selection\end{tabular} & \begin{tabular}[c]{@{}c@{}}Server-Client \\Architecture\end{tabular} & Data Cache & PyPI Install & AL Strategy Zoo     \\
        \midrule
        DeepAL \cite{deepal} &                           &                    &                            &              &            & \checkmark \\
        ModAL \cite{modal}   &                           &                    &                            &              & \checkmark & \checkmark \\
        ALiPy \cite{alipy}   &                           &                    &                            &              & \checkmark & \checkmark \\
        libact \cite{libact} &                           &                    &                            &              & \checkmark & \checkmark \\
        \textbf{ALaaS (Ours)}& \checkmark                & \checkmark         & \checkmark                 & \checkmark   & \checkmark & \checkmark \\
        \bottomrule
    \end{tabular}
    }
\end{table}

\section{System Design and Architecture}

This section first highlights our Active-Learning-as-a-Service (ALaaS) with two key features, then details the workflow and the core optimizations of the system as shown in Figure\ref{fig:system-arch}. 

\subsection{ALaaS Highlights}

We highlight two key features provided by our system, namely efficiency and automation. These features are also our design principles, leading the implementation to consider both experts (\textit{e.g.}, data scientists and machine learning engineers) and non-experts (\textit{e.g.}, customers with little domain knowledge) all the time. Besides, we list the main differences between our system and other existing tools in the table\ref{tab:open-source-tool-compare}.

\textbf{Efficiency.} Different from previous ALaaS python tools, our system is developed as a machine learning service. As a service, in addition to providing various AL strategies, our ALaaS offers efficiency to users by employing a lot of optimization technologies, including a pipeline process \cite{narayanan2019pipedream}, ML serving backend adoption\cite{trtserving}, and caching.

\textbf{Automation.} To further lower the barrier to use, an AL system should ensure that it can be utilized by non-AL experts with minimal effort and without writing much code. To this end, we design an AL agent to automate the AL strategy selection and the data selection procedures. With the AL agent, non-experts only need to input target accuracy and budget, then sit and wait for the final results.

\subsection{ALaaS Workflow}

\begin{figure*}[t]
  \centering
  \includegraphics[width=1.0\textwidth]{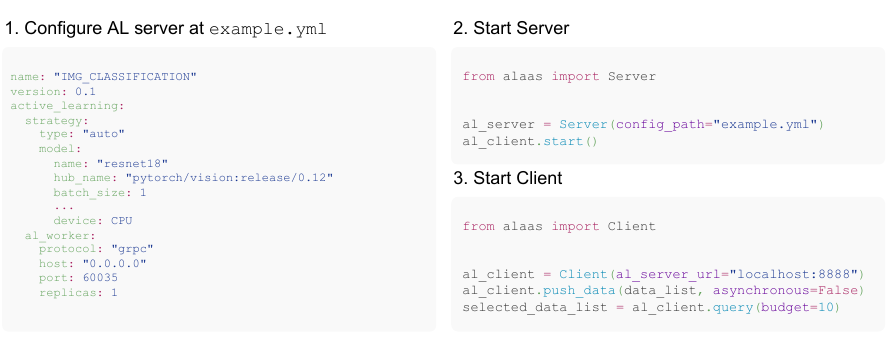}
  \caption{An AL service can be easily configured and started with \textit{YML} files.}
\label{fig:config-example}
\end{figure*}

As shown in Figure \ref{fig:system-arch}, ALaaS adopts the service-client architecture to abstract complex AL algorithms into web-based services, enabling an out-of-the-box user experience. Specifically, first, users only need to prepare a configuration file including basic settings like model name and serving device by following the provided templates, as shown in Figure \ref{fig:config-example}. Then, with very few lines of code (LoCs), users can start both the AL client and AL server. Next, users will push their unlabeled datasets to the AL server. The dataset can be stored either in the local disk or AWS S3 \cite{awss3}. Meanwhile, if users do not specify an AL strategy, ALaaS will invoke an AL agent to automatically select suitable strategies to perform data selection tasks to approach users' accuracy targets constrained by budget. Finally, the AL server will return a selected dataset for human labeling as well as further model updating. 

\subsection{ALaaS Optimization}

\begin{figure*}[]
  \centering
  \includegraphics[width=1.0\textwidth]{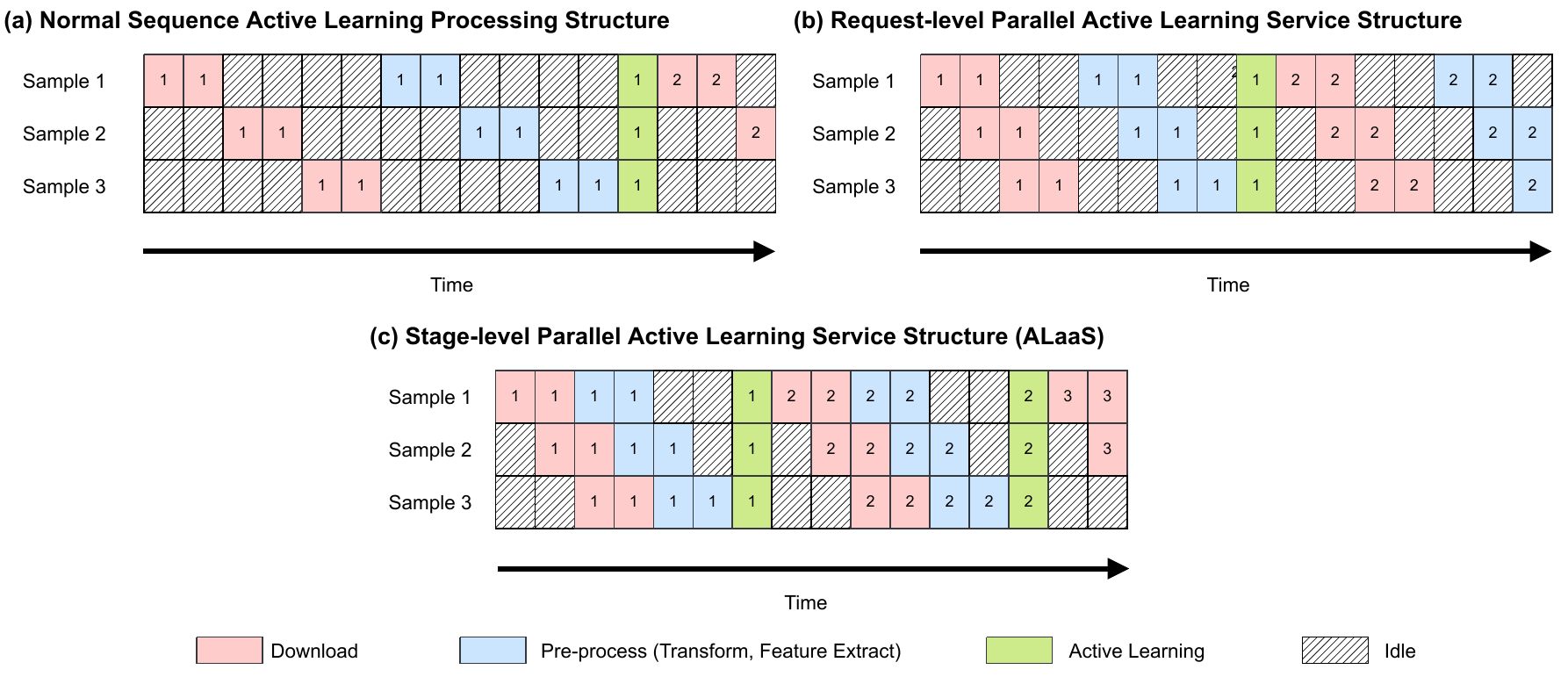}
  \caption{Dataflow comparison among conventional pool-based learning methods (a), (b), and the proposed ALaaS (c). Different colors represent different stages of the AL process, and the number inside the box indicates AL's round.}
\label{pipeline-design}
\end{figure*}

Different from previous methods only providing AL strategy for accessibility, our ALaaS achieves high efficiency and automation by implementing a pipeline processing with a data cache, and an AL agent including a performance predictor and a loop controller.

\textbf{Pipeline.} As shown in Figure\ref{pipeline-design}, we divide the processing into three stages, downloading (red color), pre-processing (blue color), and AL (green color). Once users start to send datasets, the ALaaS server will parse the datasets' Uniform Resource Identifier (URI) \cite{berners2005uniform} in the AL client and pipeline the data downloading and data pre-processing to reduce the hardware waiting time and improve the processing speed. Meanwhile, we implement a \textbf{data cache} to temporarily store the processed samples. By doing so, we further improve the processing speed. The reason is public clouds usually adopt the computation and storage separation design, and transferring the data back and forth between the current computation node to the storage node is very time-consuming. After the AL server receive a batch (batch size depends on users' settings) of raw samples, it will send them to optimized inference workers with ML serving backend for pre-processing (\textit{e.g.}, embedding calculation). Finally, those workers can return the processed data to the AL server for final data selection, then send the selected results back to the AL client. We observe that even with such a simple pipeline optimization, ALaaS's processing speed can become \textbf{10x} times faster than the other open-source platforms (see Section \ref{compare-open-source}).

\textbf{AL agent.} The AL agent implements a Predictive-based Successive Halving Early-stop (PSHEA) procedure (as shown in the Algorithm \ref{alg:early-stop}) to help users who are unsure about selecting AL strategies. Specifically, it includes a performance predictor and a loop controller. The performance predictor is a negative exponential forecasting model \cite{jin2021autolrs}. It predicts the next-round accuracy that can be achieved by a certain AL strategy under certain budget settings. The loop controller is to run many AL strategies as `candidates` first and then decide which one should be eliminated at the current round (\textit{i.e.}, an early-stopping mechanism). After several rounds of reaching pre-defined targets (\textit{e.g.}, budget or accuracy), most AL strategies will be eliminated and the left strategies with the selected data samples will be suggested to users. This design is mainly based on the conclusion that no AL strategy can outperform all others under all settings \cite{hacohen2022active}. With this AL agent, non-experts can utilize ALaaS in a cost-efficient manner rather than trying all AL strategies in a brute force manner.

\begin{algorithm}
	\caption{Predictive-based Successive Halving Early-stop (PSHEA)}
	\label{alg:early-stop}
	\begin{algorithmic}[1]
	    \State \textbf{Input:}  1. user target accuracy $a_t$; 
	    \\ \ \ \ \ \ \ \ \ \ \ \ \ 2. unlabeled data set $\xi$ with size $\tau$; 
	    \\ \ \ \ \ \ \ \ \ \ \ \ \ 3. the maximum labeling budget $b_{max} \ (b_{max} \leq \tau)$; 
	    \\ \ \ \ \ \ \ \ \ \ \ \ \ 4. active learning strategy set $L$;
	    
		\State $a_0 \gets $ pre-train the deep active learning model and get the initial evaluation accuracy.
		\State \textbf{initialize} 1. the maximum evaluation accuracy $a_{max} = a_0$; 
		\\ \ \ \ \ \ \ \ \ \ \ \ \ \ \ \ \ 2. active learning round $r = 0$; 
		\\ \ \ \ \ \ \ \ \ \ \ \ \ \ \ \ \ 3. labeled data set $d_l \gets \emptyset$ for strategy $l$; 
		\\ \ \ \ \ \ \ \ \ \ \ \ \ \ \ \ \ 4. consumed total budget $b_{total} = 0$;
		\While {\texttt{True}}
		\IIf{$a_{max} \geq a_t$} \texttt{break;} 
		\EndIIf \SComment{stop when reaches the target learning accuracy.}
		\IIf{$b_{total} \geq b_{max}$} \texttt{break;}
		\EndIIf \SComment{stop when labeling budget is not enough.}
		\IIf{\texttt{converge}} \texttt{break;} 
		\EndIIf \SComment{stop when active learning accuracy does not increase.}
		\For{strategy $l$ in $L$} \SComment{estimate the future performance of each strategy.}
		    \State $d^l \gets d^l \cup d^l_{r}$, select and label data $d^l_r$ from $\xi$ (cost budget $b^l_r$), merge labeled data into $d^l$.
            \State $a_l \gets $ update the AL model with $d^l$, get and record the evaluation accuracy $a_l$;
            \State $a_l^{*} \gets $ train a negative exponential forecasting model on historical $a_l$ and predict the
            \State \ \ \ \ \ \ \ \ \ \ active learning  accuracy $a_l^{*}$ in the next round;
            \State $b_{total} \gets b_{total} + b^l_r$ calculate the consumed total budget.
        \EndFor
        \State $r \gets r + 1$
        \State $a_{max} \gets$ update the best evaluation accuracy among all strategies;
        \If{\texttt{number of strategy in $L > 1$}} \SComment{strategy-level early-stopping}
            \State stop and remove the strategy $l^{'}$ with the least $a^{*}_{l^{'}}$ from $L$;
		\EndIf
		\EndWhile
	\end{algorithmic}
\end{algorithm}

\section{System Evaluation}

This section presents the quantitative evaluation of our systems. We first compare our system with other open-source platforms. Then we benchmark our system from different perspectives to demonstrate its efficiency and accessibility.

\subsection{Evaluation setup}
We implement our ALaaS with Python, gRPC, etc., and conduct a set of benchmark studies with the following settings.

\textbf{Hardware \& Software.} We conduct experiments on AWS EC2 and an Apple Mac Mini (M1 Chip). Our integrated data processing worker is NVIDIA Triton Inference Server \cite{trtserving}.

\textbf{Dataset.} We evaluate our system on CIFAR-10 \cite{cifar} and SVHN datasets. CIFAR-10 includes 50,000 training images and 10,000 test images. SVHN contains 600,000 images and we randomly sample 3,000 images as the test set and 10,000 images as the AL set to be labeled.

\textbf{Model.} We use the widely deployed ResNet-18 \cite{he2016deep} model to benchmark systems under many settings. We only fine-tune ResNet-18's last layer with the AL-selected and human-labeled samples.


\subsection{Comparison with other AL open source tools}
\label{compare-open-source}

The first experiment compares the efficiency of ALaaS with that of other baselines.

\textbf{Settings.}  In this experiment, we simulate a one-round AL process, which applies AL methods to scan the whole dataset to generate a sub-pool at one-time. This sub-pool includes samples that will be used to improve an existing ML model further. Specifically, we first train an ML model with randomly selected 10,000 images from the CIFAR-10 training set as the initial model. Next, we use different AL tools to serve the whole AL pipeline on an AWS 3x.large CPU/GPU EC2. For all tools, we use the same AL strategy named least confidence sampling \cite{sequential1994david}. Finally, these tools will select 10,000 samples from the rest 40,000 images in the training set and we will compare their latency and throughput for an efficiency evaluation.

\textbf{Results \& Insights.} The results are shown in Table\ref{tab:open-source-tool-perf-eval}. Compared to other tools, our ALaaS achieves the lowest latency and highest throughput while still maintaining the same updated model evaluation accuracy. This efficiency improvement can be attributed to two sides. First, our ALaaS implements stage-level parallelism which reduces the device idle time extremely. Second, ALaaS allows using multi-threads and distributed processing workers (\textit{e.g.}, Triton inference server for embedding extraction) to accelerate the data selection.

\subsection{ALaaS Characterization}

We further benchmark our ALaaS with different system settings. The first experiment is to evaluate different AL strategies re-implemented in our system. The second experiment explores the system efficiency on different batch sizes. The third experiment evaluates the proposed AL strategy auto-selection procedure with an early-stopping algorithm.

\begin{table}[]
\centering
\caption{Performance comparison among different AL open-source tools. Compared to all baselines, ALaaS has the lowest latency and the highest throughput.}
\label{tab:open-source-tool-perf-eval}
\vspace{7pt}
    \adjustbox{max width=\textwidth}{
    \begin{tabular}{lcccc}
    \toprule
    AL Open-source Tool & Top-1 Accuracy (\%) & Top-5 Accuracy (\%) & One-round AL Latency (sec)  & End-to-end Throughput (Image/sec) \\
    \midrule
    DeepAL \cite{deepal}  & 72.40               & 75.46               & 2287.00 $\pm$ 179.37      & 17.49                             \\
    ModAL \cite{modal}    & 72.40               & 75.46               & 2006.95 $\pm$ 37.98         & 19.93                           \\
    ALiPy \cite{alipy}    & 72.40               & 75.46               & 2410.85 $\pm$ 77.81         & 16.59                           \\
    libact \cite{libact}  & 71.34               & 72.32               & 1771.33 $\pm$ 109.77        & 22.58                           \\
    \textbf{ALaaS (Ours)} & 72.40               & 75.46               & \textbf{552.45 $\pm$ 30.385}        & \textbf{72.40}                           \\
    \bottomrule
    \end{tabular}
    }
\end{table}

\subsubsection{AL strategy impact}

Our ALaaS already provides many out-of-the-box AL strategies in AL Strategy Zoo for users. This experiment evaluates these strategies re-implemented by ALaaS from accuracy and efficiency views to provide more insights. All settings are the same as in the previous experiment.

\begin{figure}[h]
    \centering
    \begin{subfigure}{0.33\textwidth}
        \centering
        \includegraphics[width=1.0\linewidth]{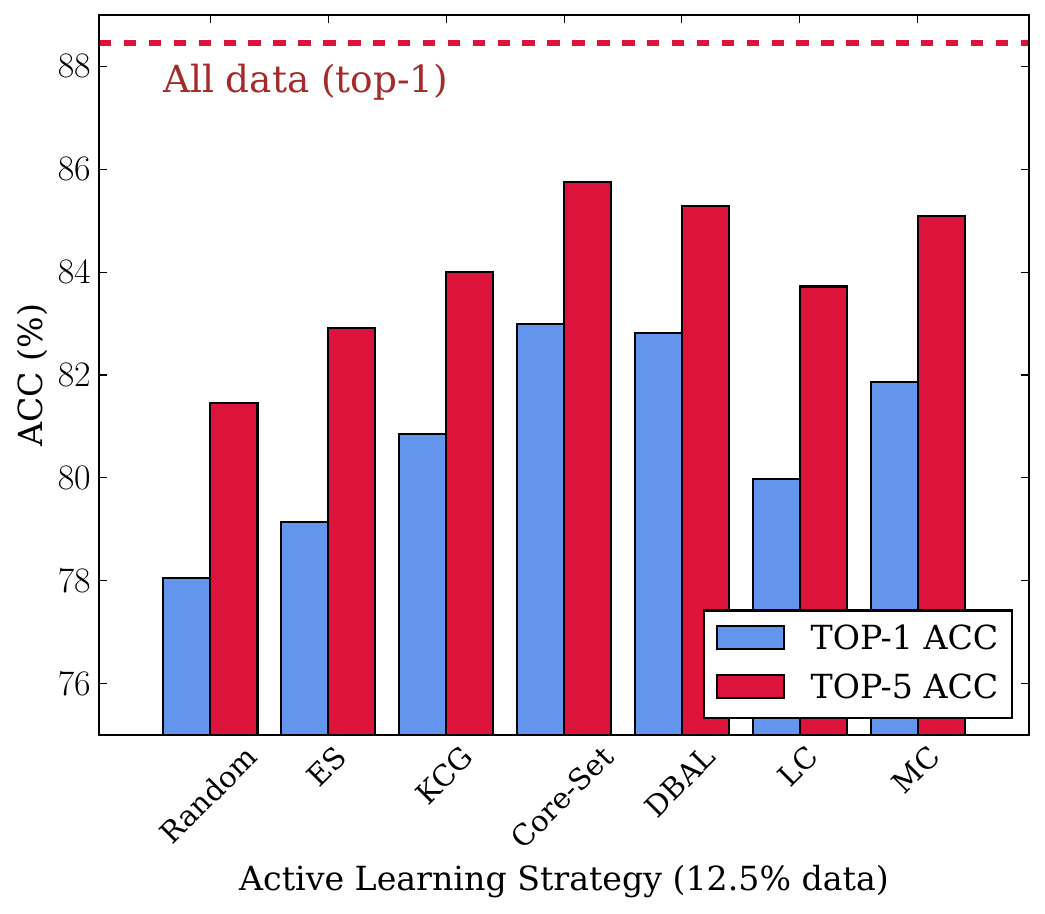}
        \caption{Accuracy of different strategies}
        \label{fig:ablation-al-strategy-acc}
    \end{subfigure}\hfill%
    \begin{subfigure}{0.33\textwidth}
        \centering
        \includegraphics[width=1.0\linewidth]{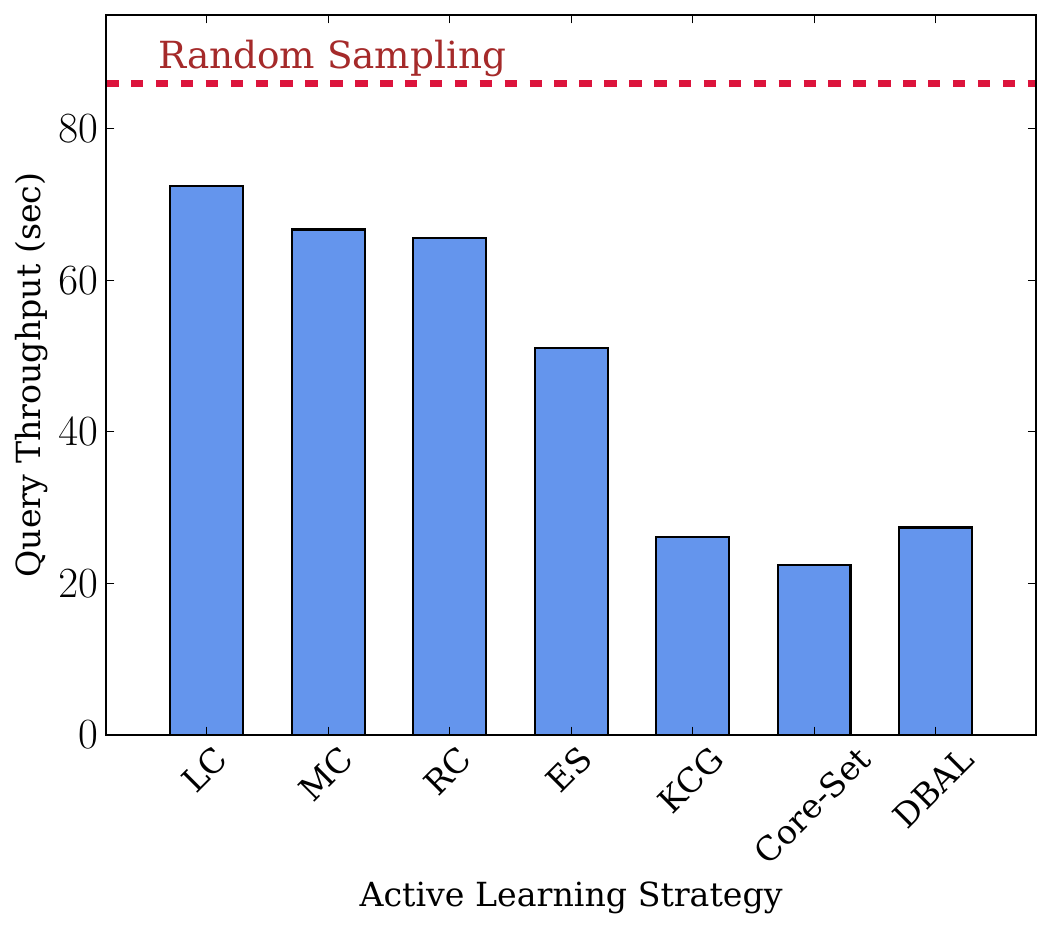}
        \caption{AL query throughput}
        \label{fig:ablation-al-strategy-throughput}
    \end{subfigure}\hfill%
    \begin{subfigure}{0.33\textwidth}
        \centering
        \includegraphics[width=1.0\linewidth]{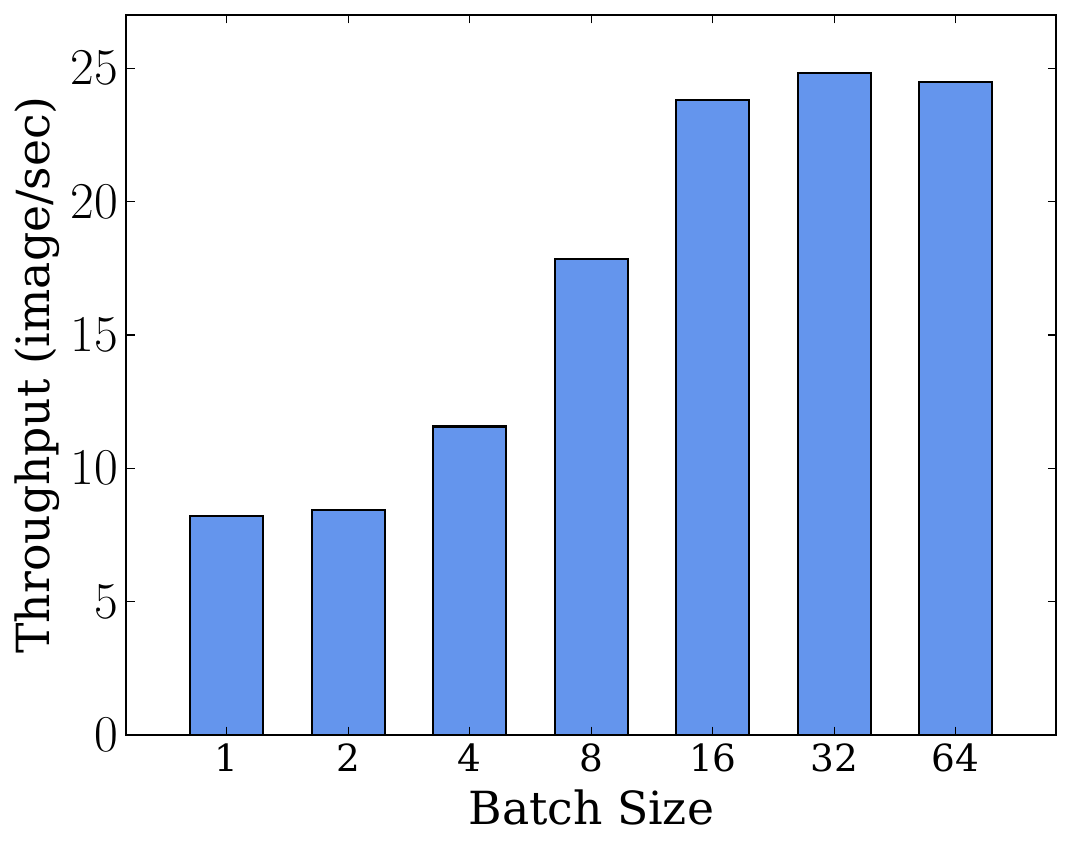}
        \caption{End-to-end throughput}
        \label{fig:ablation-al-infer-bs}
    \end{subfigure}\hfill%
    \caption{Performance of one-round AL for ResNet-18 \cite{he2016deep} on CIFAR-10 dataset \cite{cifar} using different AL strategies (\textit{i.e.}, Least Confidence (LC) \cite{sequential1994david}, Margin Confidence (MC) \cite{margconf2001tobias}, Ratio Confidence (RC) \cite{settles2009active}, Entropy Sampling (ES) \cite{settles2009active}, K-Center Greedy (KCG) \cite{alcluster2004nguyen}, Core-Set \cite{sener2018active}, and Diverse Mini-Batch (DBAL) \cite{diverse2019fedor}) (see Figures \ref{fig:ablation-al-strategy-acc}, \ref{fig:ablation-al-strategy-throughput}) and AL inference batch size (see Figure \ref{fig:ablation-al-infer-bs}). The lower-bound baseline is using random sampling (Random) strategy, while the upper-bound baseline is using the entire dataset for training.}
\label{fig:ablation-al-strategy}
\end{figure}

\textbf{Results \& Insights.} The evaluation model accuracy of different methods is shown in Figure\ref{fig:ablation-al-strategy-acc}. Core-Set \cite{sener2018active} achieves the highest accuracy with no surprise as it is designed for CNNs in Computer Vision (CV) tasks. Meanwhile, Diverse Mini-Batch (DBAL) \cite{diverse2019fedor} and Margin Confidence sampling (MC) \cite{sequential1994david} have the second and the third highest accuracy respectively, though they have proposed for a long time. This indicates that even in the deep learning (DL) era, conventional methods (\textit{e.g.}, calculating margins) still play a vital role and can cooperate with DL well.

The throughput comparison is shown in Figure\ref{fig:ablation-al-strategy-throughput}. The least confidence sampling (LC) has the highest throughput while Core-Set selection achieves the lowest throughput. Given the accuracy \ref{fig:ablation-al-strategy-acc}and the throughput \ref{fig:ablation-al-strategy-throughput} results, we can conclude that the accuracy improvement of Core-Set comes from its heavy design while conventional sampling methods like MC balances the trade-off between the accuracy and the efficiency well.

In summary, ALaaS provides many strategies with the right accuracy and can run them efficiently.

\subsubsection{Batch size impact.}

\textbf{Settings.} We evaluate the batch size (BS) impact on the public cloud (\textit{i.e.}, AWS with S3). We first store the CIFAR-10 dataset on an AWS S3 bucket. We then start ALaaS on a laptop to simulate an end-to-end AL process, including downloading data from other devices, pre-processing data, and selecting samples with an AL strategy. The other settings are the same as the first experiment.

\textbf{Results \& Insights.} Our ALaaS can manage the whole process steadily and efficiently with different batch sizes, as shown in Figure\ref{fig:ablation-al-infer-bs}. We can observe many interesting phenomena. First, BS = 1 and BS = 2 have very close throughput. Second, the increasing trend from BS = 4 to BS = 16 is dramatic. Third, after BS = 16, the increasing trend will stop. We attribute the reason that the transmission time accounts for a large proportion of the total processing time when the batch size is small. Therefore, the throughput improvement is marginal at the beginning. Then, as the batch computation time becomes the largest part of the total processing time, the improvement is dramatic. Finally, when the batch size reaches the computation capacity, this increasing diminishes.

\subsubsection{PSHEA procedure evaluation.}

\textbf{Settings.} We evaluate our AL agent with the PSHEA procedure on both CIFAR-10 and SVHN datasets. For both CIFAR-10 and SVHN experiments, we simulate an 8-round AL procedure. 

\begin{figure}[!htb]
\centering
\begin{subfigure}{0.33\textwidth}
    \centering
    \includegraphics[width=1.0\linewidth]{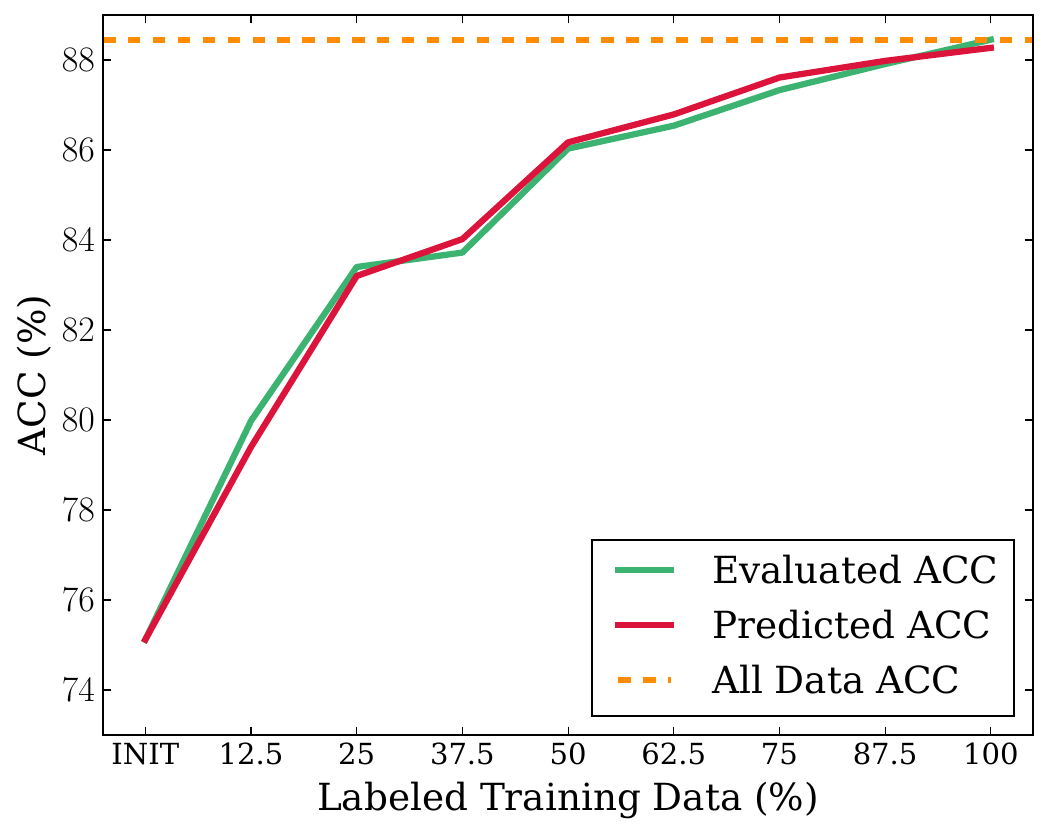}
    \caption{Accuracy prediction}
    \label{fig:ablation-al-acc-predict}
\end{subfigure}\hfill%
\begin{subfigure}{0.33\textwidth}
    \centering
    \includegraphics[width=1.0\linewidth]{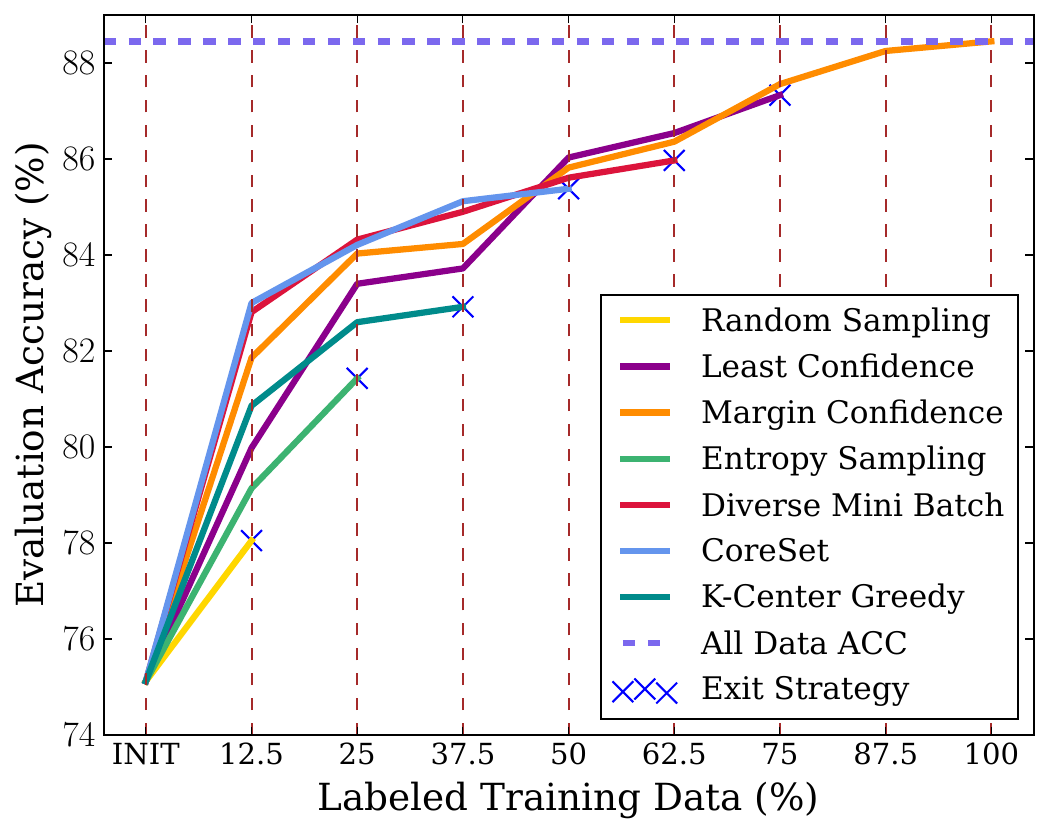}
    \caption{CIFAR-10 dataset}
    \label{fig:ablation-al-cifar-shea}
\end{subfigure}\hfill%
\begin{subfigure}{0.33\textwidth}
    \centering
    \includegraphics[width=1.0\linewidth]{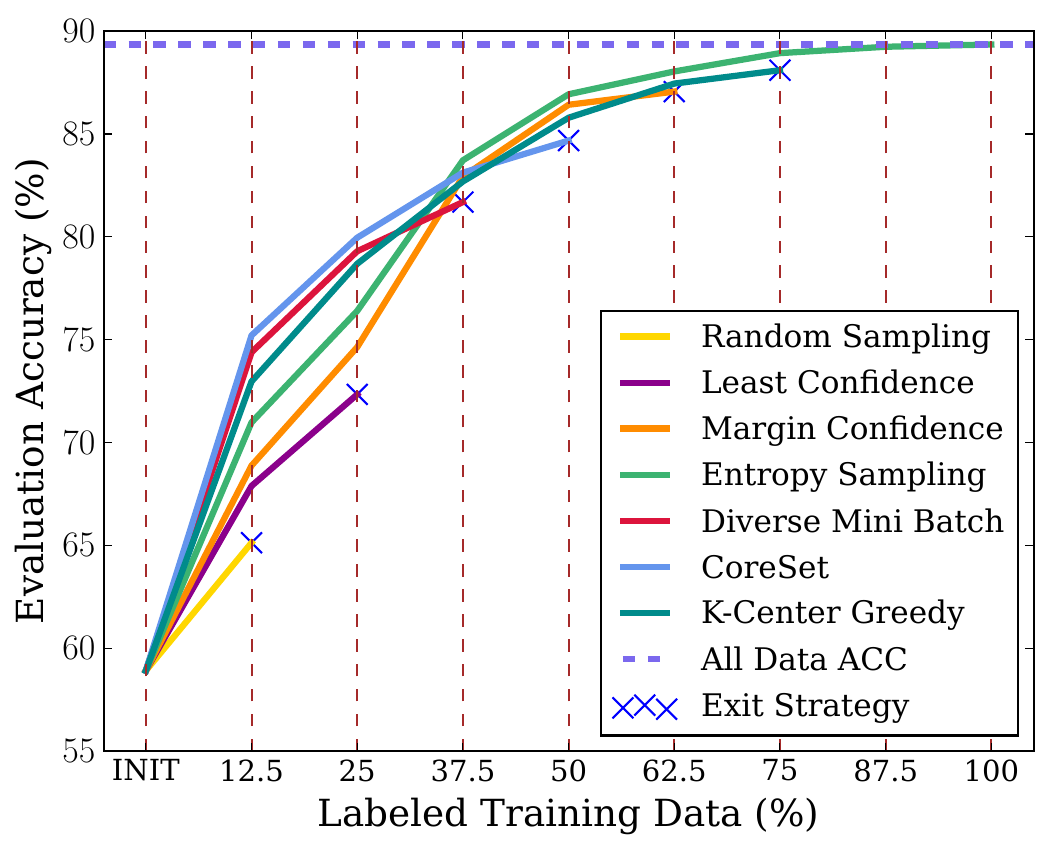}
    \caption{SVHN dataset}
    \label{fig:ablation-al-svhn-shea}
\end{subfigure}\hfill%
\caption{PSHEA prediction accuracy and multi-round results on two datasets.}
\end{figure}

\textbf{Results \& Insights.} We first use the least confidence sampling strategy to evaluate the effectiveness of our accuracy prediction model mentioned in the Algorithm\ref{alg:early-stop}. From Figure\ref{fig:ablation-al-acc-predict}, we observe that the prediction model can foresee the accuracy very accurately, laying a solid foundation for later AL strategy auto-selection. Then, we run our AL agent with PSHEA. The AL agent will launch all 7 AL strategies as candidates in the first round and then eliminate strategies round by round. As shown in Figure\ref{fig:ablation-al-cifar-shea}, for two different datasets, our AL agent selects different AL strategies under different budges (\textit{i.e.}, the percentage of the labeled data in this experiment). This observation aligns with the conclusion from previous studies \cite{hacohen2022active} and further shows the necessity of multi-round auto-selection, as no AL method can always out perform others under different budgets and datasets. Our method early stops many of the AL selection processes and helps to save running costs.

\section{Conclusion}
This paper presents a new MLOps system, named ALaaS, for data-centric AI. ALaaS adopts the philosophy of Machine-Learning-as-Service and implements a server-client architecture, so users can use AL as a web service. Meanwhile, our system employs stage-level parallelism, data cache, and batching to improve AL running efficiency. Furthermore, it includes an AL agent with a predictive-based successive halving early-stop (PSHEA) procedure to select suitable AL strategies for users under different budgets and accuracy targets. Experiments show that our system has lower latency and higher throughput compared to all other baselines. Experiments also show that the auto-selection process can help users eliminate low-performing AL strategies earlier and save cost. We release our code at GitHub to facilitate the AL research.



\newpage
\bibliography{nips_ref.bib}
\bibliographystyle{plainnat}










\end{document}